\documentclass[sigconf]{acmart}
\AtBeginDocument{%
  }

\copyrightyear{2025}
\acmYear{2025}
\setcopyright{cc}
\setcctype{by}
\acmConference[MMSports '25]{Proceedings of the 8th International ACM Workshop on Multimedia Content Analysis in Sports}{October 27--28, 2025}{Dublin, Ireland}
\acmBooktitle{Proceedings of the 8th International ACM Workshop on Multimedia Content Analysis in Sports (MMSports '25), October 27--28, 2025,
Dublin, Ireland}\acmDOI{10.1145/3728423.3759416}
\acmISBN{979-8-4007-1835-9/2025/10}

\acmSubmissionID{mspt43}

\usepackage{enumitem}
\usepackage{pifont}
\usepackage{multirow}

\usepackage{xcolor}
\usepackage[table]{xcolor}
\definecolor{color1}{rgb}{0.65,0.95,0.95}
\definecolor{color2}{rgb}{0.7, 0.85, 1}
\definecolor{color3}{rgb}{0.7, 0.7, 1}

\begin{document}

\title{GTATrack: Winner Solution to SoccerTrack 2025 with Deep-EIoU and Global Tracklet Association}

\author{Rong-Lin Jian}
\affiliation{
 \institution{Institute of Intelligent Systems\\College of AI, National Yang Ming Chiao Tung University}
 \city{Tainan}
 \country{Taiwan}
}
\author{Ming-Chi Luo}
\affiliation{
 \institution{Department of Computer Science\\, National Yang Ming Chiao Tung University}
 \city{Hsinchu}
 \country{Taiwan}
}
\author{Chen-Wei Huang}
\affiliation{
 \institution{Master Program in Remote Sensing Science and Technology  \\ National Central University}
 \city{Taoyuan}
 \country{Taiwan}
}

\author{Chia-Ming Lee}
\affiliation{
 \institution{Institute of Data Science\\National Cheng Kung University}
 \city{Tainan}
 \country{Taiwan}
}
\author{Yu-Fan Lin}
\affiliation{
 \institution{Miin Wu School of Computing\\National Cheng Kung University}
 \city{Tainan}
 \country{Taiwan}
}
\author{Chih-Chung Hsu}
\affiliation{
 \institution{Institute of Intelligent Systems\\College of AI, National Yang Ming Chiao Tung University}
 \city{Tainan}
 \country{Taiwan}
}

\renewcommand{\shortauthors}{Rong-Lin Jian et al.}

\renewcommand{\shortauthors}{Rong-Lin Jian et al.}

\begin{abstract}
Multi-object tracking (MOT) in sports is highly challenging due to irregular player motion, uniform appearances, and frequent occlusions. These difficulties are further exacerbated by the geometric distortion and extreme scale variation introduced by static fisheye cameras. In this work, we present \textbf{GTATrack}, a hierarchical tracking framework that win first place in the SoccerTrack Challenge 2025. GTATrack integrates two core components: \textit{Deep Expansion IoU (Deep-EIoU)} for motion-agnostic online association and \textit{Global Tracklet Association (GTA)} for trajectory-level refinement. This two-stage design enables both robust short-term matching and long-term identity consistency. Additionally, a pseudo-labeling strategy is used to boost detector recall on small and distorted targets. The synergy between local association and global reasoning effectively addresses identity switches, occlusions, and tracking fragmentation. Our method achieved a winning HOTA score of 0.60 and significantly reduced false positives to 982, demonstrating state-of-the-art accuracy in fisheye-based soccer tracking. Our code is available at \hyperlink{https://github.com/ron941/GTATrack-STC2025}{https://github.com/ron941/GTATrack-STC2025}.
\end{abstract}




\maketitle

\section{Introduction}

\begin{figure*}
		\centering
		\includegraphics[width=0.85\textwidth]{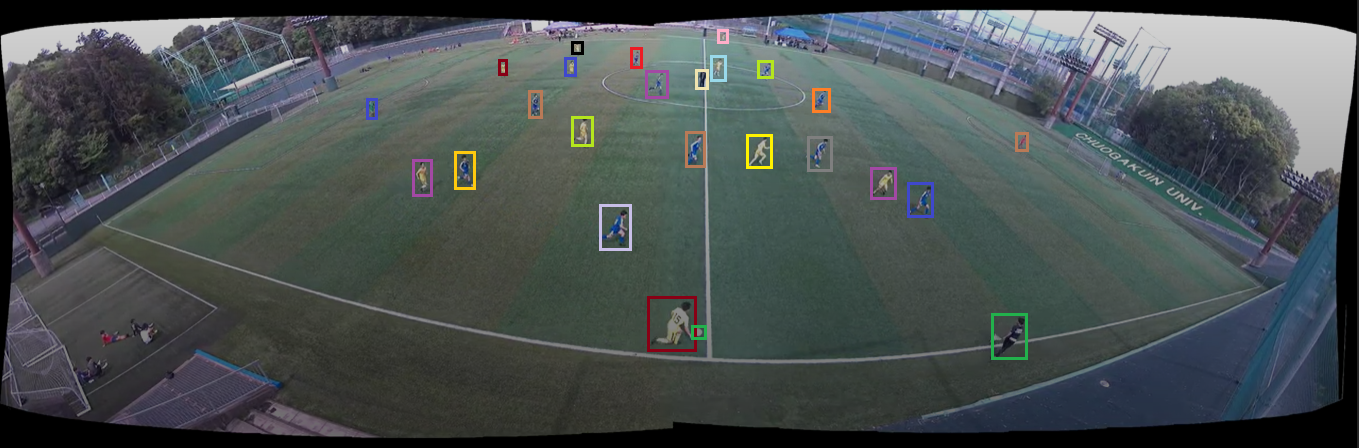}
		\caption{An overview of the complex tracking environment in a fisheye camera view. This frame illustrates several key challenges simultaneously: extreme scale variation between distant and nearby players, unpredictable spatial distribution reflecting irregular motion, and noticeable geometric distortion in peripheral areas. A robust tracker must effectively handle all these issues to maintain high accuracy.}
		\label{fig:hardcase2}	
\end{figure*}

Multi-object tracking (MOT), the task of identifying and following multiple object trajectories over time, is a fundamental problem in computer vision. While substantial progress has been made in tracking common targets such as pedestrians and vehicles~\cite{Bergmann_2019,SORT,bytetrack}, extending MOT to dynamic sports environments introduces a significantly more complex set of challenges. Unlike conventional scenarios characterized by predictable, linear motion, players in sports exhibit highly irregular movement patterns with abrupt changes in direction and rapid acceleration~\cite{sportsMOT,Cioppa_2022,soccernet,soccernet_v2}. This is further complicated by severe appearance similarity due to uniform jerseys, frequent and prolonged occlusions from physical contact, and the use of static fisheye cameras that introduce geometric distortions, extreme scale variations, and resolution degradation for distant players~\cite{scott2024teamtrackdatasetmultisportmultiobject,scott2022soccertrack,broadtrack}.

These challenges collectively undermine the reliability of conventional MOT pipelines. Kalman filter-based methods~\cite{SORT,Kalman}, which assume linear and smooth motion, fail to capture the abrupt, non-linear trajectories common in sports, leading to fragmented tracks and frequent identity switches. Detection-focused approaches~\cite{yolox,so-detr} often miss small or distorted players, especially in peripheral regions of fisheye views with extreme scale variation. Appearance-based methods~\cite{deepsort,SOLIDER-ReID} also struggle under uniform jerseys, motion blur, and lighting inconsistencies, making identity association highly ambiguous.

Crucially, these issues are interdependent: weak detections increase the burden on appearance features, and poor ReID embeddings further degrade trajectory linking. Combined with fisheye-induced distortion, these limitations propagate through the tracking pipeline, resulting in unstable and error-prone performance. Addressing these challenges demands a unified framework that handles motion irregularity, scale-aware detection, appearance ambiguity, and long-term identity consistency in a coherent way.

To address these challenges, we present \textbf{GTATrack}, a unified two-stage tracking framework that bridges local spatial association and global temporal refinement. GTATrack begins with \textbf{Deep-EIoU}~\cite{deepeiou}, a motion-agnostic online tracker that bypasses motion prediction in favor of iterative bounding box expansion and deep ReID features~\cite{osnet-reid,SOLIDER-ReID}, enabling robust tracking under erratic motion. On top of this, we apply \textbf{GTA-Link}~\cite{gta} as a global post-processing module that performs trajectory-level reasoning to merge fragmented tracks and resolve identity switches caused by occlusions or appearance ambiguity~\cite{wang2021splitconnectuniversaltracklet,strongsort,translink}.

Our key insight is that robust tracking in sports demands complementary reasoning across multiple temporal scales. This hierarchical formulation proved highly effective in the \textbf{SoccerTrack Challenge 2025}, where GTATrack achieved \textbf{first place} with a leading HOTA score of 0.60, significantly outperforming strong baselines.
%\vspace{4pt}
Our main contributions are summarized as follows:
\begin{itemize}
    \item \textbf{GTATrack: A hierarchical local-global tracking framework} that combines motion-agnostic online association with global trajectory refinement for stable and consistent tracking in fisheye soccer videos.
    
    \item \textbf{Deep-EIoU for robust local association}, which utilizes iterative bounding box expansion and deep appearance features to handle irregular motion without relying on predictive models.
    
    \item \textbf{GTA-Link for global trajectory refinement}, which resolves long-term identity switches by clustering fragmented tracklets using spatio-temporal and appearance-based reasoning.
\end{itemize}

The remainder of this paper is organized as follows. Section~\ref{sec:relatedworks} reviews related works in object detection, sports MOT, and person ReID. Section~\ref{sec:motivation} elaborates on our motivation. Section~\ref{sec:methodology} details our proposed hierarchical tracking framework. Section~\ref{sec:experiment} presents the experimental setup, ablation studies, and final challenge results. Finally, Section~\ref{sec:conclusion} concludes the paper.

\begin{figure*}
		\centering
		\includegraphics[width=0.85\textwidth]{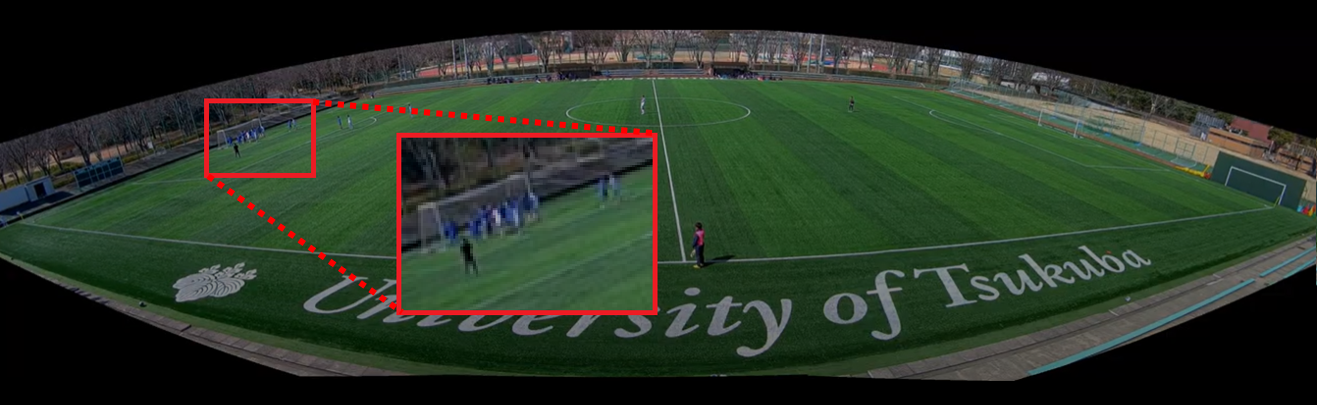}
		\caption{Illustration of tracking challenges in non-central regions of a fisheye camera view. The goalmouth area, a critical zone for gameplay, suffers from severe geometric distortion. Players in this region appear small and are often densely clustered, leading to frequent and prolonged occlusions. %These compounding factors represent a significant hurdle for robust detection and tracking.
        }
		\label{fig:hardcase1}	
        \vspace{-0.3cm}
\end{figure*}

\section{Related Works}
\label{sec:relatedworks}

\subsection{Object Detection for Fisheye Camera}

\textbf{Scale-aware Object Detection.} A significant challenge in using fixed fisheye cameras for sports tracking is the extreme scale variation across the playing field~\cite{FRIDA, SPinSVMC, RAPiD}. Players near the camera appear large, while those at a distance become extremely small and suffer from insufficient resolution, demanding a detector with strong multi-scale capabilities. Modern architectures, such as the YOLO~\cite{yolox,yolov11} series, address this through improved backbone and neck designs that enhance multi-scale feature extraction and fusion. To further improve recall on the smallest, low-resolution targets, specialized detectors like SO-DETR~\cite{so-detr} can be utilized, and fine-tuning model parameters, such as redefining anchor box sizes based on the dataset's bounding box distribution, is a common and effective practice.

\textbf{Non-rigid Object Distortion.} Detection is further complicated by distortions from two sources \cite{10222872,Gochoo_2023_CVPR,10678572}. First, fisheye lenses introduce significant geometric distortions that can warp the appearance of players, particularly near the edges of the image~\cite{sport1,drct}. Second, athletes themselves are highly non-rigid objects whose posture and aspect ratio change dramatically during play, for instance, when a player is standing upright in one frame and lying on the ground in the next. To ensure robustness against these shape variations, it is crucial to train detectors on augmented data \cite{fisheye1,fisheye2} that simulate diverse poses and to make practical adjustments, such as removing strict aspect ratio constraints, to prevent the model from filtering out valid but unconventionally shaped targets.

\subsection{Multi-Object Tracking in Sport-scenes}

\textbf{Traditional Paradigm and Recent Developments.} Multi-object tracking (MOT) in sports typically follows the dominant Tracking-by-Detection paradigm, where targets are first detected in each frame and then associated into trajectories \cite{scott2024teamtrackdatasetmultisportmultiobject,scott2022soccertrack,sportsMOT,Giancola_2022,sset}. However, this approach faces significant hurdles in sports scenarios that are less prevalent in general pedestrian tracking. Athletes exhibit highly irregular motion, and near-identical team uniforms severely impede appearance-based discrimination, leading to frequent identity confusion. Moreover, intense physical contact and dense player formations result in severe and prolonged occlusions. These challenges have spurred recent developments and the creation of specialized benchmarks. For instance, the frequent re-entry of players in soccer matches, a key issue highlighted by datasets like SoccerNet-Tracking~\cite{Cioppa_2022}, poses a critical test for a tracker's long-term ReID capabilities.

\textbf{Irregular Motion.} A primary challenge in sports MOT is handling irregular motion. Traditional online trackers, such as SORT~\cite{SORT} and DeepSORT~\cite{deepsort} , rely on the Kalman Filter~\cite{Kalman}  for motion prediction. However, the filter's underlying assumption of linear motion is frequently violated by the erratic movements of athletes, resulting in poor prediction accuracy and subsequent tracking failures. While some methods like OC-SORT~\cite{oc-sort} have sought to refine this predictive model, a more fundamental shift is seen in approaches like Deep-EIoU \cite{deepeiou}. Deep-EIoU \cite{deepeiou} discards motion prediction entirely, instead performing direct data association using an Iterative Scale-Up ExpansionIoU mechanism combined with deep appearance features~\cite{deepeiou}. This motion-agnostic strategy provides inherent robustness against unpredictable movements, proving highly effective at maintaining tracking continuity in dynamic sports environments.
\captionsetup{skip=10pt}
\begin{table}[h!]
\centering
\small
\caption{Comparative overview of Sport-related datasets.}
\label{tab:dataset_comparison}
\scalebox{0.95}{
\begin{tabular}{l r r l}
\hline
\hline
\textbf{Dataset} & \textbf{\#Frames} & \textbf{\#BBoxes} & \textbf{Domain} \\
\midrule
SSET \cite{sset} & 12,000 & 12,000 & Soccer \\
\midrule
SN-Tracking \cite{Giancola_2022} & 225,375 & 3,645,661 & Soccer \\
\midrule
SportsMOT \cite{sportsMOT} & 150,379 & 1,629,490 & \begin{tabular}{@{}l@{}}Soccer\\Basketball\\Volleyball\end{tabular} \\
\midrule
SoccerTrack V1 \cite{scott2022soccertrack} & 82,800 & 2,484,000 & Soccer \\
\midrule
TeamTrack \cite{scott2024teamtrackdatasetmultisportmultiobject}& \textbf{279,900} & \textbf{4,374,900} & \begin{tabular}{@{}l@{}}Soccer\\Basketball\\Handball\end{tabular} \\
\midrule
\textbf{SoccerTrack V2} & 35,932 & 795,054 & Soccer\\
\hline
\hline
\end{tabular}}
\end{table}

\subsection{Person Re-Identification in Sport Scenes}

Person re-identification (ReID) serves as a core component in appearance-based multi-object tracking, especially in sports scenarios with high occlusion frequency, fast motion, and low inter-class variance due to uniform team attire \cite{8099840,9025455}. Unlike general pedestrian tracking, distinguishing athletes with similar appearances under distortion and scale variation is considerably more difficult \cite{10656857,scott2024teamtrackdatasetmultisportmultiobject}.

Recent ReID models have focused on enhancing discriminative power through multi-scale design or long-range attention. OSNet~\cite{osnet-reid} learns omni-scale features to capture both fine-grained textures and global body structures, proving effective under viewpoint variation. Transformer-based architectures, such as SOLIDER~\cite{SOLIDER-ReID} and TransLink~\cite{translink}, introduce global context modeling and temporal attention to improve identity continuity. For fisheye or non-central views, robust embeddings remain critical. Prior work such as surround-view pedestrian ReID~\cite{zhao2021fisheyereid,scott2022soccertrack} further highlights the need for geometry-invariant features in wide-angle scenes.

Despite these efforts, maintaining identity consistency in sport-specific MOT remains an open challenge. The compounded factors of camera distortion, repetitive appearance, and dense formations demand lightweight yet expressive ReID models tailored to occlusion resilience and high-speed tracking systems.

\section{Motivation}
\label{sec:motivation}

\begin{figure*}
		\centering
		\includegraphics[width=1\textwidth]{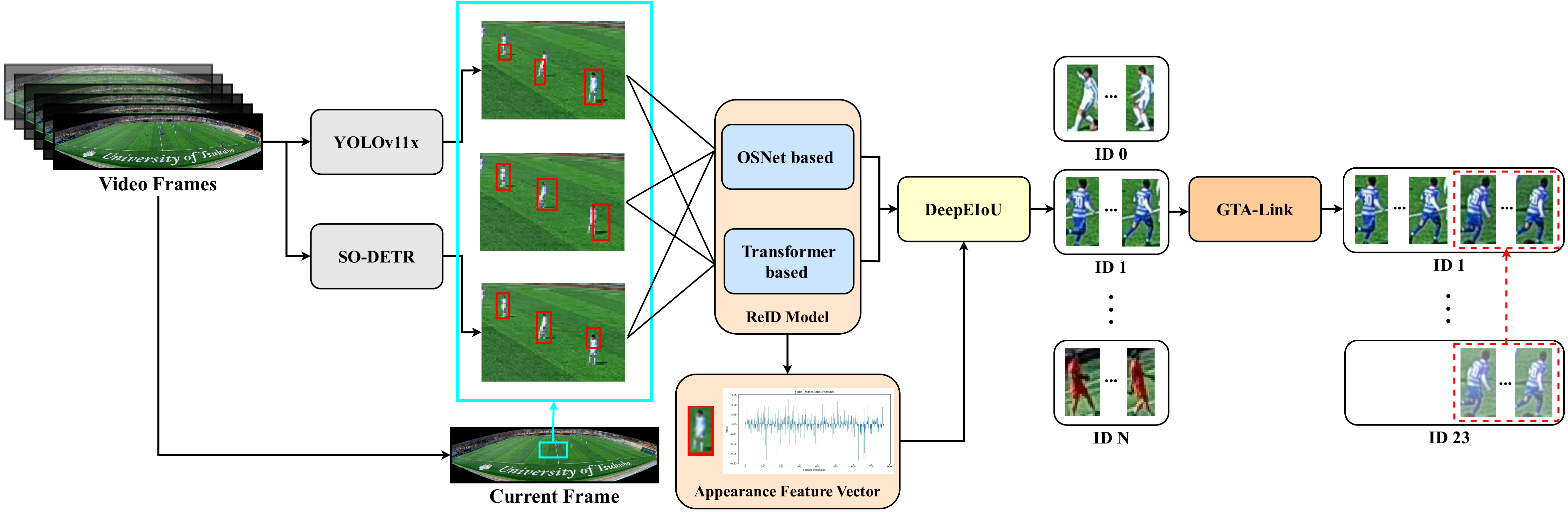}
	   \caption{Overview of our GTATrack framework. (1) Object detection with a detector (e.g., YOLOv11x~\cite{yolov11}). (2) Appearance feature extraction using a ReID model (e.g., OSNet~\cite{osnet-reid}). (3) Online tracking via Deep-EIoU~\cite{deepeiou} to form initial tracklets. (4) Offline refinement with GTA-Link~\cite{gta} to merge fragments and correct identity switches.}

		\label{fig:framework}	
\end{figure*}
Multi-object tracking in sports poses greater challenges than conventional scenarios due to irregular motion, frequent occlusions, and high appearance similarity from uniformed players. These difficulties are further exacerbated in the SoccerTrack Challenge 2025, where static fisheye cameras introduce geometric distortion, extreme scale variation, and reduced resolution—conditions that often lead to fragmented trajectories and identity switches.

To address these compounded challenges, we propose a hierarchical framework that combines local association and global refinement. Specifically, we integrate Deep-EIoU~\cite{deepeiou}, a motion-agnostic online tracker, with GTA-Link~\cite{gta}, a trajectory-level global association module. This two-stage design allows for robust short-term matching and long-term identity consistency, even under severe fisheye-induced distortion. The effectiveness of this approach is demonstrated by our first-place performance in the SoccerTrack Challenge 2025.

\section{Methodology}
\label{sec:methodology}
\subsection{Overview}

Our proposed framework, \textbf{GTATrack}, adopts a modular two-stage design that integrates motion-agnostic online tracking with global offline refinement. The pipeline consists of four components:

\begin{itemize}
    \item \textbf{Object Detection}: A YOLOv11x detector locates players in each frame, with pseudo-labeling used to enhance recall on small or distant targets.
    
    \item \textbf{Online Tracking}: Deep-EIoU performs frame-to-frame association using a cost function that combines Expansion IoU and ReID-based appearance similarity.
    
    \item \textbf{Re-Identification}: OSNet extracts L2-normalized appearance features to ensure identity consistency under occlusion and motion blur.
    
    \item \textbf{Offline Refinement}: GTA-Link merges fragmented tracklets via hierarchical clustering based on appearance and spatio-temporal constraints.
\end{itemize}

\subsection{Object Detection}

The first stage of our pipeline focuses on detecting all players in each frame. Given an input frame $I_t$, the detector $\mathcal{D}$ outputs a set of bounding box detections $O_t$:

\begin{equation}
    O_t = \mathcal{D}(I_t) = \{o_1, o_2, \dots, o_k\}
    \label{eq:detection_output}
\end{equation}

Each detection $o_i \in O_t$ consists of a bounding box $b_i = (x_i, y_i, w_i, h_i)$ and a confidence score $s_i$:

\begin{equation}
    o_i = (b_i, s_i)
    \label{eq:detection_element}
\end{equation}

We considered two detection architectures for $\mathcal{D}$: the convolutional one-stage detector YOLOv11x~\cite{yolov11}, and SO-DETR~\cite{so-detr}, a Transformer-based model tailored for small object detection. Both models were pretrained and fine-tuned on task-specific soccer data. Due to its efficiency and strong performance across varying scales, particularly in handling small and distant players common in fisheye views, YOLOv11x was selected as the final detector.

\textbf{Pseudo-labeling Strategy.}  
To further improve recall, we employed a semi-supervised learning strategy based on pseudo-labeling. The fine-tuned YOLOv11x model was used to generate predictions on unlabeled video frames, and high-confidence detections were retained as pseudo-labels. These pseudo-labeled instances were then incorporated into the training set, allowing the model to benefit from a more diverse data distribution and enhanced representation of challenging cases.

\begin{figure}
		\centering
		\includegraphics[width=0.48\textwidth]{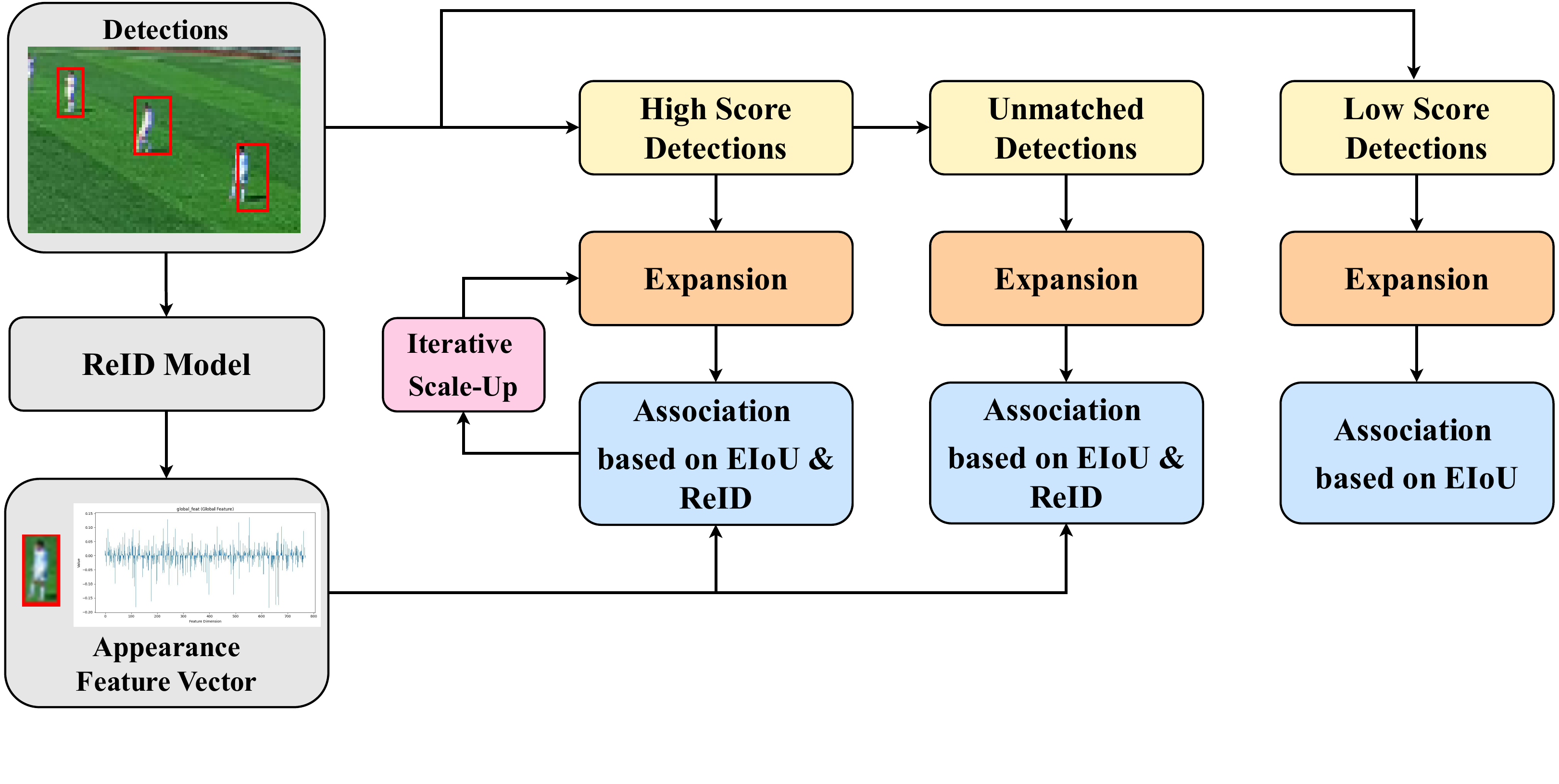}
        \caption{Deep-EIoU~\cite{deepeiou} adopts a multi-stage matching strategy that combines Expansion IoU and ReID features to prioritize high-quality associations. An iterative scale-up mechanism expands the spatial search area, improving robustness to occlusion and irregular motion.}

		\label{fig:deepeiou}	
\end{figure}

\subsection{Person Re-Identification}

To maintain consistent identities across frames despite occlusions and appearance variations, we employ a person ReID module that serves as a feature extractor $\mathcal{R}$. For each cropped player image patch $c_i$, the module generates a $D$-dimensional feature vector:

\begin{equation}
    f_i = \mathcal{R}(c_i)
    \label{eq:feature_extraction}
\end{equation}

Each feature vector is L2-normalized, ensuring $\|f_i\|_2 = 1$, which enables efficient computation of appearance similarity using cosine distance. The pairwise appearance distance between two objects $i$ and $j$ is defined as:

\begin{equation}
    d_{\text{app}}(i, j) = 1 - \text{sim}(f_i, f_j) = 1 - (f_i \cdot f_j)
    \label{eq:appearance_distance}
\end{equation}

We considered two architectures for $\mathcal{R}$: the lightweight CNN-based OSNet~\cite{osnet-reid}, and the Transformer-based SOLIDER~\cite{SOLIDER-ReID}. OSNet is optimized for multi-scale feature extraction, making it effective in capturing both local details and global semantics, which is beneficial under fast motion and frequent occlusions. Due to its compact design and consistent representation quality, OSNet was selected as the ReID backbone in our system.

\subsection{Online Tracking}

For frame-to-frame association, we adopt \textbf{Deep-EIoU}~\cite{deepeiou} as our core online tracking algorithm. The task is formulated as a linear assignment problem between the set of current detections $O_t$ and existing tracklets $\mathcal{T}$. We compute a cost matrix $\mathbf{C}$, where each element $C_{ij}$ reflects the matching cost between detection $o_i$ and tracklet $\tau_j$, based on both spatial and appearance cues.

The spatial cost is defined using the Expansion IoU (EIoU) between bounding boxes:

\begin{equation}
    C_{\text{spatial}}(i, j) = 1 - \text{EIoU}(b_i, b_{\tau_j})
    \label{eq:spatial_cost}
\end{equation}

where $b_i$ and $b_{\tau_j}$ denote the bounding boxes of the detection and the latest tracklet state. The appearance cost $C_{\text{app}}(i, j)$ is computed as the cosine-based distance $d_{\text{app}}(i, j)$ introduced in Equation~\ref{eq:appearance_distance}. The final assignment cost is the weighted sum of these two terms.

To determine the optimal matching, we minimize the total association cost:

\begin{equation}
    \min_{\mathbf{X}} \sum_{i} \sum_{j} C_{ij} X_{ij}
    \label{eq:assignment_problem}
\end{equation}

where $X_{ij} = 1$ indicates that detection $o_i$ is assigned to tracklet $\tau_j$, and $0$ otherwise. The assignment is efficiently solved using the Hungarian algorithm~\cite{kuhn1955hungarian}.

Unlike traditional Kalman filter-based trackers~\cite{Kalman}, Deep-EIoU does not rely on motion prediction. Instead, it employs a motion-agnostic strategy that leverages bounding box expansion and appearance features, making it more robust to abrupt direction changes and irregular trajectories common in sports scenarios.

\begin{figure}[ht]
    \centering
    \includegraphics[width=0.95\linewidth, trim=10 10 10 10, clip]{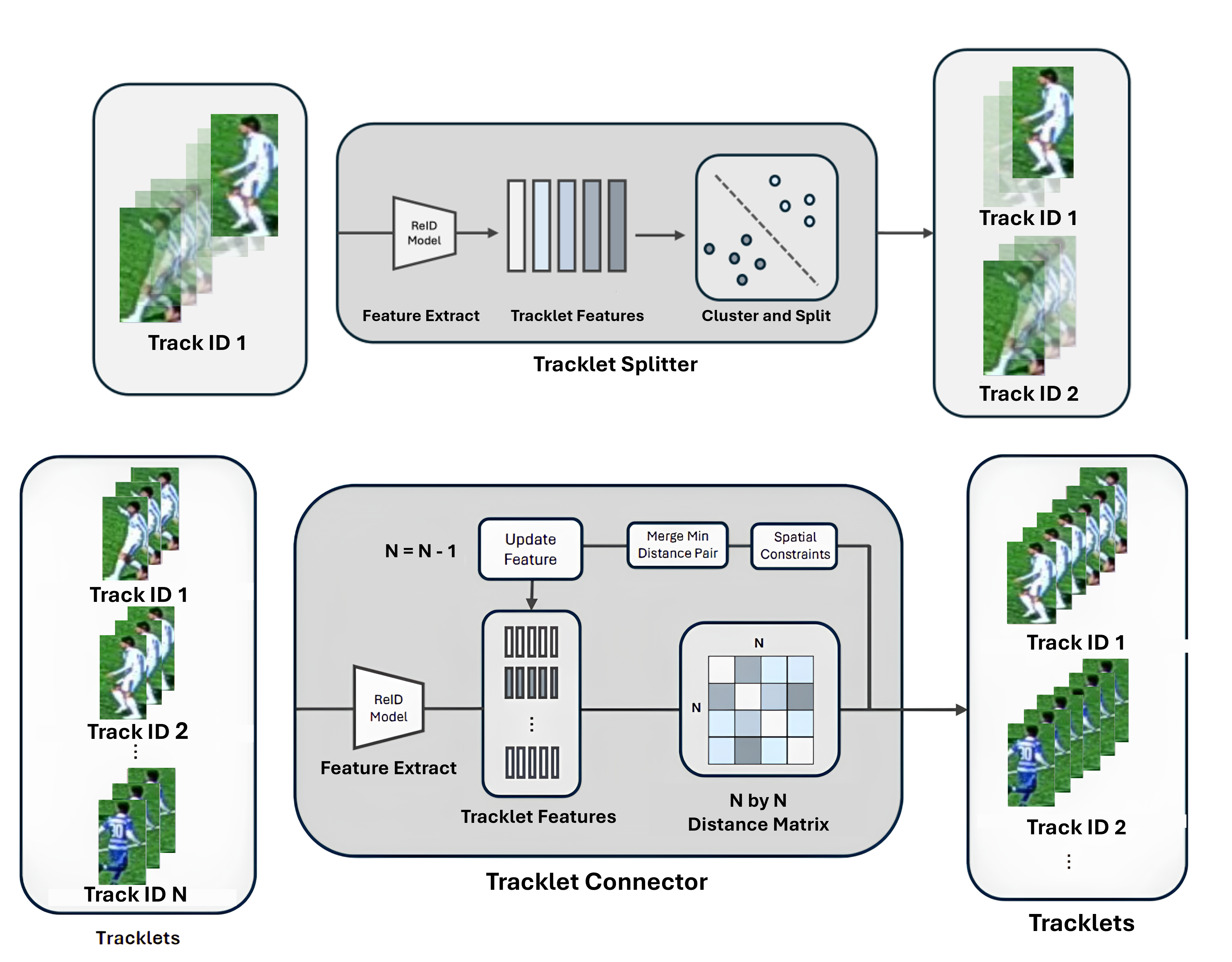}
    \caption{GTA-Link~\cite{gta} includes a Splitter for identity separation and a Connector for trajectory merging via spatio-temporal and appearance cues.}
    \label{fig:gta}
\end{figure}

\subsection{Offline Refinement}

To address identity switches and fragmented trajectories from the online stage, we apply \textbf{GTA-Link}~\cite{gta} as a global post-processing module. GTA-Link operates on the set of initial tracklets $\mathcal{T}_{\text{initial}}$, treating each as a node in a graph and performing trajectory-level association through appearance-based clustering.

The core component, the \emph{Tracklet Connector}, merges fragmented tracklets using hierarchical clustering based on pairwise appearance distance. The distance between two tracklets $\tau_i$ and $\tau_j$ is computed as the average pairwise distance across all feature embeddings:

\begin{equation}
    D_{\text{app}}(\tau_i, \tau_j) = \frac{1}{L_i L_j} \sum_{m=1}^{L_i} \sum_{n=1}^{L_j} d_{\text{app}}(f_{i,m}, f_{j,n})
    \label{eq:tracklet_distance}
\end{equation}

where $L_i$ and $L_j$ are the lengths of tracklets $\tau_i$ and $\tau_j$, and $d_{\text{app}}$ is the cosine-based appearance distance defined in Equation~\ref{eq:appearance_distance}. Tracklet pairs with the lowest $D_{\text{app}}$ are merged iteratively, subject to spatial and temporal constraints and a similarity threshold $\alpha$.

GTA-Link also includes a \emph{Tracklet Splitter} module, which identifies and separates multiple identities within a single tracklet by clustering inconsistent appearance features. In this work, we focus on the Connector module to consolidate trajectories and enhance long-term identity consistency.

\section{Experiment Results}
\label{sec:experiment}

\subsection{Experiment Setup}

\textbf{Dataset and Splitting.} Our experiments are conducted on the official dataset from the SoccerTrack Challenge 2025. This dataset features soccer matches recorded by static fisheye cameras, which introduce significant challenges, including severe geometric distortion~\cite{fisheye1,fisheye2}, extreme scale variations, low resolution in distant areas, and frequent occlusions due to high player density and similar uniforms.
\begin{table}[h]
\centering
\small
\begin{tabular}{lccccc}
\hline
\hline
\textbf{\#Video} & \textbf{Resolution} & \textbf{\#Frames} & \textbf{\#Boxes} & \textbf{\#IDs} & \textbf{Split} \\
\midrule
118577 & \multirow{6}{*}{4096$\times$1080} & 6000 & 132,000 & \multirow{6}{*}{22} & Train \\
118576 &  & 5975  & 131,450 &  & Train \\
118575 &  & 6000  & 132,000 &  & Train \\
117092 &  & 6000  & 132,000 &  & Train \\
117093 &  & 5976  & 131,472 &  & Test \\
128058 &  & 5981  & 131,582 &  & Test \\
\hline
\hline
\end{tabular}
\caption{Statistics of the six SoccerTrack training videos. All videos are captured at 4096×1080 resolution with 22 unique player identities. A subset is used as validation for ablation studies.}
\label{tab:dataset-stat}
\vspace{-0.5cm}
\end{table}

\begin{table*}[h!]
\centering
\small
\caption{Comprehensive ablation study on the validation set. We evaluate different framework configurations to validate our choice of online tracker (Deep-EIoU~\cite{deepeiou}), the necessity of offline refinement (GTA-Link~\cite{gta}), and the impact of the detector's training strategy. Best results are highlighted as \colorbox{color1}{1st}, \colorbox{color2}{2nd}.}
\label{tab:ablation_main}
\begin{tabular}{l | c c c c c c c}
\hline
\hline
\textbf{Framework Configuration} & \textbf{HOTA ↑} & \textbf{IDSW ↓} & \textbf{LocA ↑} & \textbf{DetA ↑} & \textbf{AssA ↑} & \textbf{FN ↓} & \textbf{FP ↓} \\
\hline
ByteTrack~\cite{bytetrack} (baseline) & 0.42 & 630 & 0.84 & 0.74 & 0.26 & 10115.0 & 961.0 \\
Deep-EIoU~\cite{deepeiou} & \cellcolor{color2}{0.54} & \cellcolor{color2}{325.5} & 0.84 & \cellcolor{color1}{0.76} & \cellcolor{color2}{0.38} & \cellcolor{color1}{5440.5} & \cellcolor{color1}{980.0} \\
Deep-EIoU~\cite{deepeiou}+GTA-Link~\cite{gta} (with Finetuning) & \cellcolor{color1}{0.60} & \cellcolor{color1}{331.5} & 0.84 & \cellcolor{color1}{0.76} & \cellcolor{color1}{0.47} & \cellcolor{color2}{5454.5} & \cellcolor{color2}{982.0} \\
\hline
\hline
\end{tabular}\vspace{-0.2cm}
\end{table*}

For our experiments, we partitioned the provided videos into a training set (4 videos) and a validation set (2 videos) for ablation studies. All sequences were captured at a resolution of 4096×1080 and contain 22 unique player identities. Detailed statistics for each video are provided in Table~\ref{tab:dataset-stat}.

%To simulate realistic application scenarios, we selected four videos for training and two for testing.

%\vspace{0.5em}
\textbf{Implementation Details.}
All experiments were conducted on a single NVIDIA RTX 3090 GPU. Our models were trained for 200 epochs using the AdamW optimizer with a learning rate of 0.0001, employing multi-scale and mosaic data augmentations for robustness. The primary detector, YOLOv11x~\cite{yolov11}, was configured with an input resolution of 1280 pixels on the longer side and a batch size of 12. Our two-stage tracking framework first utilizes Deep-EIoU~\cite{deepeiou} for online association, matching detections via iterative spatial expansion and OSNet-based~\cite{osnet-reid} appearance features. Subsequently, GTA-Link~\cite{gta} performs offline refinement, using global association to merge fragmented trajectories and enhance long-term identity consistency.
%\begin{itemize}
 %   \item \textbf{Detection (YOLOv11x)}: Resolution = 1920, batch size = 12, optimizer = AdamW. Multi-scale and mosaic augmentations are enabled.
   % \item \textbf{Detection (SO-DETR)}: Transformer-based small-object detector used in ensemble evaluation.
    %\item \textbf{Tracking (Deep-EIoU)}: Uses iterative spatial expansion + ReID embedding (OSNet / Transformer).
    %\item \textbf{Tracking (GTA-Link)}: Post-processing module for tracklet-level global identity association.
   % \item \textbf{Training}: 200 epochs, 1× RTX 3090 GPU, learning rate = 0.0001.
%\end{itemize}

\vspace{0.5em}
\subsection{Evaluation Metric}
We evaluate tracking performance using the following standard metrics:

\textbf{HOTA (Higher Order Tracking Accuracy)} is the primary metric, measuring the geometric mean of detection and association accuracy~\cite{hota}:
\begin{equation}
\text{HOTA}_\alpha = \sqrt{\text{DetA}_\alpha \cdot \text{AssA}_\alpha}
\end{equation}
Final HOTA, DetA, AssA, and LocA scores are averaged over all thresholds $\alpha$.

\textbf{IDSW (Identity Switches)} counts the number of times a predicted identity switches to another. Lower values indicate better identity consistency.

\textbf{AssA (Association Accuracy)} reflects the quality of trajectory linking, computed as the average Jaccard Index between ground-truth and predicted trajectories:
\begin{equation}
\text{AssA}_\alpha = \frac{1}{|TP_\alpha|} \sum_{c \in TP_\alpha} \mathcal{A}(c)
\end{equation}

\textbf{LocA (Localization Accuracy)} measures the spatial precision of matches via average IoU:
\begin{equation}
\text{LocA}_\alpha = \frac{1}{|TP_\alpha|} \sum_{c \in TP_\alpha} \text{IoU}(c)
\end{equation}

\textbf{DetA (Detection Accuracy)} is the Jaccard Index over matched, missed (FN), and false positive (FP) detections:
\begin{equation}
\text{DetA}_\alpha = \frac{|TP_\alpha|}{|TP_\alpha| + |FN_\alpha| + |FP_\alpha|}
\end{equation}

\subsection{Experiment Results}

To validate our architectural design, we conducted an ablation study on the validation set, evaluating the impact of each major component, including the online tracker, offline refinement, and pseudo-labeling strategy (Table~\ref{tab:ablation_main}).

Replacing a conventional motion-based tracker~\cite{bytetrack} with the motion-agnostic Deep-EIoU~\cite{deepeiou} led to a notable HOTA improvement (0.52 to 0.55), primarily due to an increase in Association Accuracy (0.38 to 0.42), confirming the advantage of modeling irregular motion. The integration of GTA-Link~\cite{gta} with pseudo-labeled detector training further improved tracking performance, boosting HOTA to 0.60 and significantly reducing false positives. These results underscore the effectiveness of our two-stage framework in addressing identity switches and maintaining long-term trajectory consistency.

\subsection{Ablation Studies}

\subsubsection{Impact of Pseudo Labels.}
To investigate the role of pseudo labels, we compare the performance with and without their inclusion during detector finetuning. As shown in Table~\ref{tab:ablation_study}, applying pseudo labels results in a notable HOTA increase to 0.511 and a significant reduction in false positives. The pseudo labels were derived from the model's predictions on the official training set, followed by the selection of high-confidence samples for fine-tuning. This semi-automatic process effectively enhances the detector's ability to recall challenging small-scale players, especially in distant or distorted regions.

\begin{table}[h!]
\centering
% \caption{Ablation experiment with and without manual frame marking}
\caption{Ablation study for pseudo label. Incorporating pseudo labels for fine-tuning significantly improves model performance, increasing HOTA and substantially reducing false positives and false negatives.
}
\label{tab:ablation_study}
\scalebox{0.9}{
\begin{tabular}{c c c c c}
\hline
\hline
\textbf{Method} & \textbf{Configuration} & \textbf{HOTA $\uparrow$} & \textbf{FP $\downarrow$} & \textbf{FN $\downarrow$} \\
\midrule
\multirow{2}{*}{GTATrack}&w/o pseudo label & 0.380 & 4913 & 40046 \\
&w/ pseudo label & \textbf{0.491} & \textbf{494} & \textbf{16186} \\
\hline
\hline
\end{tabular}}
\vspace{-0.3cm}
\end{table}

\begin{table*}[ht!]
\centering
%\small
\caption{SoccerTrack Challenge 2025 Leaderboard Top 5 Results.}
\label{tab:challenge_results}
\begin{tabular}{c| l |c c c c c c c}
\hline
\hline
\textbf{\#} & \textbf{Team Name} & \textbf{HOTA $\uparrow$} & \textbf{IDSW $\downarrow$} & \textbf{LocA $\uparrow$} & \textbf{DetA $\uparrow$} & \textbf{AssA $\uparrow$} & \textbf{FN $\downarrow$} & \textbf{FP $\downarrow$} \\
\hline
\textbf{1} & \textbf{wccjs (Ours)} & \textbf{0.60 (1)} & 331.50 (12) & 0.84 (13) & 0.76 (6) & 0.47 (3) & 5454.50 (6) & 982.00 (2) \\
2 & mdk-tdu & 0.59 (2) & 143.00 (1)  & 0.86 (8) & 0.77 (3) & 0.45 (4) & 5129.50 (4) & 3285.00 (11) \\
3 & SShota & 0.59 (3) & 170.00 (2)  & 0.86 (7) & 0.77 (4) & 0.45 (5) & 6090.50 (8) & 3176.50 (9) \\
4 & takashun13 & 0.59 (4) & 271.00 (8)  & 0.83 (17) & 0.74 (11) & 0.47 (2) & 4711.50 (3) & 2623.50 (6) \\
5 & YMori22 & 0.59 (5) & 267.50 (7)  & 0.83 (18) & 0.73 (12) & 0.47 (1) & 4703.00 (2) & 2823.00 (7) \\
\hline
\hline
\end{tabular}
\end{table*}
\subsubsection{Impact of Different Detector and ReID Model Combinations}

\begin{table}[h!]
\centering
\small
\caption{Ablation study of different combinations of detectors and ReID models using the Deep-EIoU~\cite{deepeiou} tracking backbone.}
\label{tab:reid_detector_ablation}
\scalebox{0.9}{
\begin{tabular}{c|c|c|c}
\hline
\hline
\textbf{Method} & \textbf{ReID Model} & \textbf{Detector} & \textbf{HOTA $\uparrow$} \\
\midrule
\multirow{4}{*}{GTATrack}&\multirow{2}{*}{OSNet~\cite{osnet-reid}} 
 & YOLOv11x~\cite{yolov11} & \textbf{0.491} \\ 
 
 && SO-DETR~\cite{so-detr} & 0.405 \\ 
 \cline{2-4}
&\multirow{2}{*}{SOLIDER~\cite{SOLIDER-ReID}} 
 & YOLOv11x~\cite{yolov11} & 0.474 \\ 
 && SO-DETR~\cite{so-detr} & 0.357 \\ 
\hline
\hline
\end{tabular}}
\end{table}

We evaluated four combinations of detector and ReID modules using a fixed Deep-EIoU~\cite{deepeiou} tracking backbone. As shown in Table~\ref{tab:reid_detector_ablation}, the combination of YOLOv11x~\cite{yolov11} and OSNet~\cite{osnet-reid} achieved the best performance with a HOTA score of 0.511, while the worst configuration—SO-DETR~\cite{SOLIDER-ReID} with Transformer-based ReID—yielded only 0.357. Due to its superior accuracy, especially on small and distant players, YOLOv11x was selected as our final detector, replacing more memory-intensive alternatives such as SO-DETR~\cite{so-detr}.

For ReID, OSNet~\cite{osnet-reid} consistently outperformed the Transformer-based model across all detector pairings. Its lightweight design and strong multi-scale feature extraction made it more resilient to occlusions and fast player motion. In contrast, Transformer-based ReID exhibited instability under high-motion conditions, leading to more identity switches. We therefore adopted YOLOv11x + OSNet as the final configuration, forming a stable foundation for our Deep-EIoU and GTA-Link tracking pipeline.

\begin{table}[h!]
\centering
\small
\caption{HOTA scores and Tracklets counts under different GTA-Link~\cite{gta}hyperparameters. Each cell shows HOTA / Tracklets after Splitter / Connector. The original Tracklets count without GTA-Link~\cite{gta} is 53.}
\label{tab:gtalink_combined}
\scalebox{0.9}{
\begin{tabular}{c|ccc}
\hline\hline
\textbf{Min-samples$\backslash$Eps} & \textbf{0.3} & \textbf{0.5} & \textbf{0.7} \\
\hline
\textbf{7}  & 0.557 / 131 / 38 & \textbf{0.568} / 69 / 27 & 0.564 / 53 / 27 \\
\textbf{10} & 0.548 / 131 / 38 & 0.565 / 65 / 27 & 0.564 / 53 / 27 \\
\textbf{13} & 0.542 / 129 / 43 & \textbf{0.568} / 66 / 29 & 0.564 / 57 / 27 \\
\hline\hline
\end{tabular}}
\vspace{-0.3cm}
\end{table}

\begin{figure}
		\centering
		\includegraphics[width=0.45\textwidth]{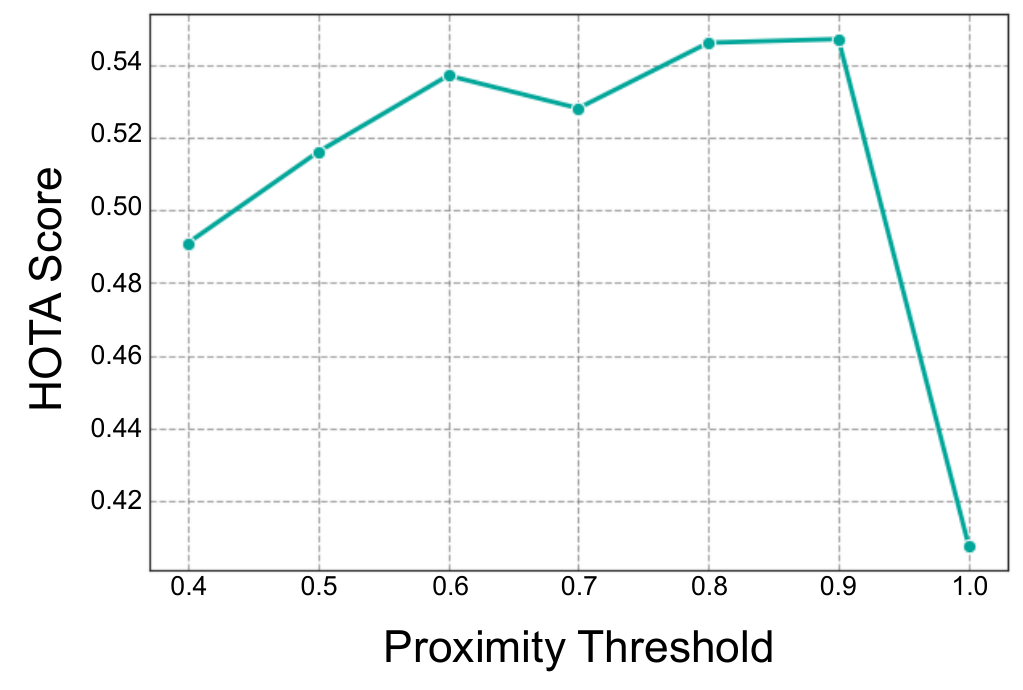}
		\caption{HOTA results under different Deep-EIoU \cite{deepeiou} proximity thresholds, where the 0.4 threshold result is obtained from a model fine-tuned using pseudo labels to enhance tracking performance.}
		\label{fig:deepeiou_hyperparam}	
\end{figure}
\subsubsection{Hyperparameter Analysis}

We analyzed the impact of the proximity threshold in Deep-EIoU~\cite{deepeiou} and found that increasing it from 0.4 to 0.9 improves HOTA from 0.491 to 0.547 (Figure~\ref{fig:deepeiou_hyperparam}), an 11.4\% gain. The best results occur in the 0.8–0.9 range, indicating that moderately relaxed spatial constraints enhance robustness when tracking small or fast-moving players. In contrast, setting the threshold to 1.0 removes spatial filtering entirely, causing mismatches between visually similar but spatially distant players, especially when propagated through GTA-Link~\cite{gta} merging.

Additionally, our GTA-Link~\cite{gta} refinement module boosts identity consistency by 3\%–4\% in HOTA across different parameter settings (Table~\ref{tab:gtalink_combined}). The combination of Connector and Splitter reduces redundant tracklets and improves association quality. For instance, under \texttt{eps} = 0.5 and \texttt{min-samples} = 7, the system split 53 tracklets into 69 fragments and merged them into 27 refined trajectories. However, the Splitter's benefit is scenario-dependent and may cause over-fragmentation in stable scenes. An adaptive activation strategy could further enhance robustness.

\subsection{Challenge Results}

Our proposed framework, \textbf{GTATrack}, ranked \textbf{first} in the \textbf{SoccerTrack Challenge 2025}, achieving the highest \textbf{HOTA score of 0.60} under the challenging fisheye setting. As shown in Table~\ref{tab:challenge_results}, GTATrack outperformed all competing entries with a superior balance between detection and association accuracy.

Notably, our method produced only 982 false positives—substantially lower than other top submissions—thanks to our pseudo-label-enhanced detector. While some competitors reached similar association scores, they suffered from unstable detection or fragmented tracks. GTATrack's motion-agnostic tracking and global refinement ensured robust identity continuity even under severe occlusions and visual ambiguity.

\section{Conclusion}
\label{sec:conclusion}

We presented \textbf{GTATrack}, a hierarchical multi-object tracking framework designed for the unique challenges of fisheye soccer videos. Our method achieved \textbf{first place} in the SoccerTrack Challenge 2025 with a leading \textbf{HOTA score of 0.60}, demonstrating its effectiveness in handling irregular motion, severe occlusion, and extreme scale variation. GTATrack combines three key components: (1) a YOLOv11x detector enhanced via a pseudo-labeling strategy that significantly improves recall on small, distant players and reduces false positives by nearly 90\%; (2) Deep-EIoU, a motion-agnostic online tracker that eliminates reliance on predictive models for robust short-term association; and (3) GTA-Link, a global post-processing module that refines trajectories through identity-aware merging. Extensive experiments confirm that this two-stage design—bridging local association and global reasoning—offers a robust and scalable solution for high-fidelity tracking in complex sports environments.

\bibliographystyle{ACM-Reference-Format}
\bibliography{mmsport_ref}

% \appendix

\end{document}